# Video-based Formative and Summative Assessment of Surgical Tasks using Deep Learning

**Erim Yanik[1], Uwe Kruger[1], Xavier Intes[1], Rahul Rahul[1], and Suvranu De[1*]**

**To ensure satisfactory clinical outcomes, surgical skill assessment must be objective, time-efficient, and preferentially automated — none of which is currently achievable. Video-based assessment (VBA) is being deployed in intraoperative and simulation settings to evaluate technical skill execution. However, VBA remains manually- and time-intensive and prone to subjective interpretation and poor inter-rater reliability. Herein, we propose a deep learning (DL) model that can automatically and objectively provide a high-stakes summative assessment of surgical skill execution based on video feeds and low-stakes formative assessment to guide surgical skill acquisition. Formative assessment is generated using heatmaps of visual features that correlate with surgical performance. Hence, the DL model paves the way to the quantitative and reproducible evaluation of surgical tasks from videos with the potential for broad dissemination in surgical training, certification, and credentialing.**

The skill of the surgeon is the single most important determinant of the success of a surgical procedure[1]. Assessment of surgical skills may be formative or summative. Formative assessment is low-stakes. Experts typically provide it as guidance during surgery. On the other hand, summative assessment is employed in high-stake certification or credentialing and is usually associated with a quantitative score computed by proctors. Though direct observation of surgeons in the operating room or on a simulator remains the current gold standard of surgical skill evaluation, video-based assessment (VBA) is receiving increasing attention[2–4]. The American Board of Surgery (ABS) is exploring VBA as a component of the Continuous Certification Program for general surgeons and related specialties[5]. However, as a post-hoc procedure, VBA is manual- and time-intensive, subjective, and prone to poor inter-rater reliability[2,3]. Moreover, VBA methodologies often entail editing the videos into snippets to reduce the workload[3], promoting subjectivity due to the editor's bias[2,3]. Further, numerous studies have reported inferior validity evidence and inflated score prediction via edited videos compared with complete videos[3]. Another limitation is that VBA is almost exclusively formative, i.e., low-stakes, and there is a notable gap in the literature regarding the use of VBA for summative, i.e., high-stakes, assessment[3], such as Fundamentals of Laparoscopic Surgery (FLS), a prerequisite for board certification in general surgery and ob/GYN surgery[6]. Hence, there is a need to develop an objective, efficient and automated approach for VBA.

Several deep learning (DL) models have been developed for automated and objective skill assessment[7], most of which rely on obtaining sensor-based kinematics data from surgeons, which is time- and labor-intensive and may interfere with the surgical task. In contrast, videos are collected routinely as part of most surgical procedures[8], making large-scale data collection feasible. Existing video-based DL models utilize editing to simplify the problem[9,10]. In addition, these models use label-preserving snippeting in which each snippet shares the label of the complete video. This is problematic as labels for the entire video may not apply to an individual snippet. Finally, current DL models do not provide means to assess the salient features characterizing performance. Explainable artificial intelligence (XAI) techniques[11,12], such as class activation maps (CAMs)[13], have not been shown to provide formative evaluation reliably.

[1]Center for Modeling, Simulation, and Imaging for Medicine (CeMSIM), Rensselaer Polytechnic Institute, USA.
*e-mail: des@rpi.edu

To address these limitations, we propose a DL model, the VBA-Net, that can utilize complete surgical video sequences to provide summative surgical scores and generate formative feedback correlated with surgical performance. Fig, 1 illustrates the overview of the study. Two datasets involving surgical pattern cutting (PC), one of the five tasks of the FLS certification program, were collected (Fig. 1a). Specifically, the *primary pattern cutting dataset* was used to develop the model, i.e., select the hyperparameters. In addition, an independently collected *additional pattern cutting dataset* was used to validate the model's generalizability on unseen subjects. Finally, to further elucidate the generalizability of the VBA-Net, we benchmarked it against the high-performing models on the *JIGSAWS dataset*[14].

VBA-Net consists of two stages (Fig. 1b). First, we utilize Mask Region-based Convolutional Neural Network (Mask R-CNN) to extract tool motion sequences from the PC videos. Second, these sequences are fed into a denoising autoencoder (DAE) for embedding used by a classifier to predict the summative scores, i.e., FLS scores and binary skill classes, and provide formative visual feedback via CAMs. For the JIGSAWS comparison, we directly use the provided kinematics in the DAE. Finally, we present a model-agnostic tool to validate the CAMs.

## Results

**Datasets**. PC entails laparoscopic scissors to cut a circular pattern printed on a 10 cm × 10 cm gauze pad while applying traction with the Maryland Dissector (grasper). The performance scores in FLS are high-stakes based on end-point metrics, e.g., time and precision error[15]. These scores categorize subjects into pass/fail classes (Extended Data Table 1) based on a cut-off threshold[15].

The JIGSAWS dataset[14], on the other hand, contains sensor-based data collected via the da Vinci Surgical System (Intuitive Surgical, Inc) for tasks: suturing, needling passing, and knot tying[14]. The dataset has three surgical skill classes, viz., novice (N), intermediate (I), and expert (E), based on the hours spent in the operating room (OR). Moreover, modified Objective Structured Assessment of Technical Skills (OSATS) scores are available. OSATS is a formative assessment rubric[3] computed based on low-stakes informative criteria[14,16]. In addition, global rating scales (GRS), a summation of individual items in the rubric, is also offered.

**Performance of Mask R-CNN.** Mask R-CNN was trained using frames from the PC datasets. As a result, it successfully extracted bounding box centroids (X, Y) from the surrounding artifacts, e.g., mechanical clips, in challenging conditions such as overlapping tools and blurred frames (Extended Data Fig. 1), reporting an average precision of 0.97 when the intersection over union (IoU) is 0.5. Here, IoU is the overlap ratio between the ground truth and the predicted bounding boxes, and 0.5 is the threshold for true prediction[17-20].

**Performance on the primary PC dataset.** The DAE and the classifier were trained via the motion sequences using the stratified 10-Fold cross-validation (CV) scheme. After extracting the salient features, the classifier robustly correlated (Fig. 2a.) the ground truth and the predicted FLS scores with an average $\rho$ of $0.915 \pm .002$ after ten training sessions with p<0.05 for each session.

| | Dataset | Population | Sample Size | Data | Scores | Skill Class |
|---|---|---|---|---|---|---|
| Pattern Cutting | Primary | 21 Students | 2055 | Videos | FLS – Summative | Pass/Fail |
| | Additional | 12 Students | 307 | | | |
| | JIGSAWS | 8 Surgeons | 39/28/36* | Kinematics | OSATS – Formative | E/I/N** |

*suturing/needle passing/knot tying, **E/I/N: expert, intermediate, novice

**(a)**

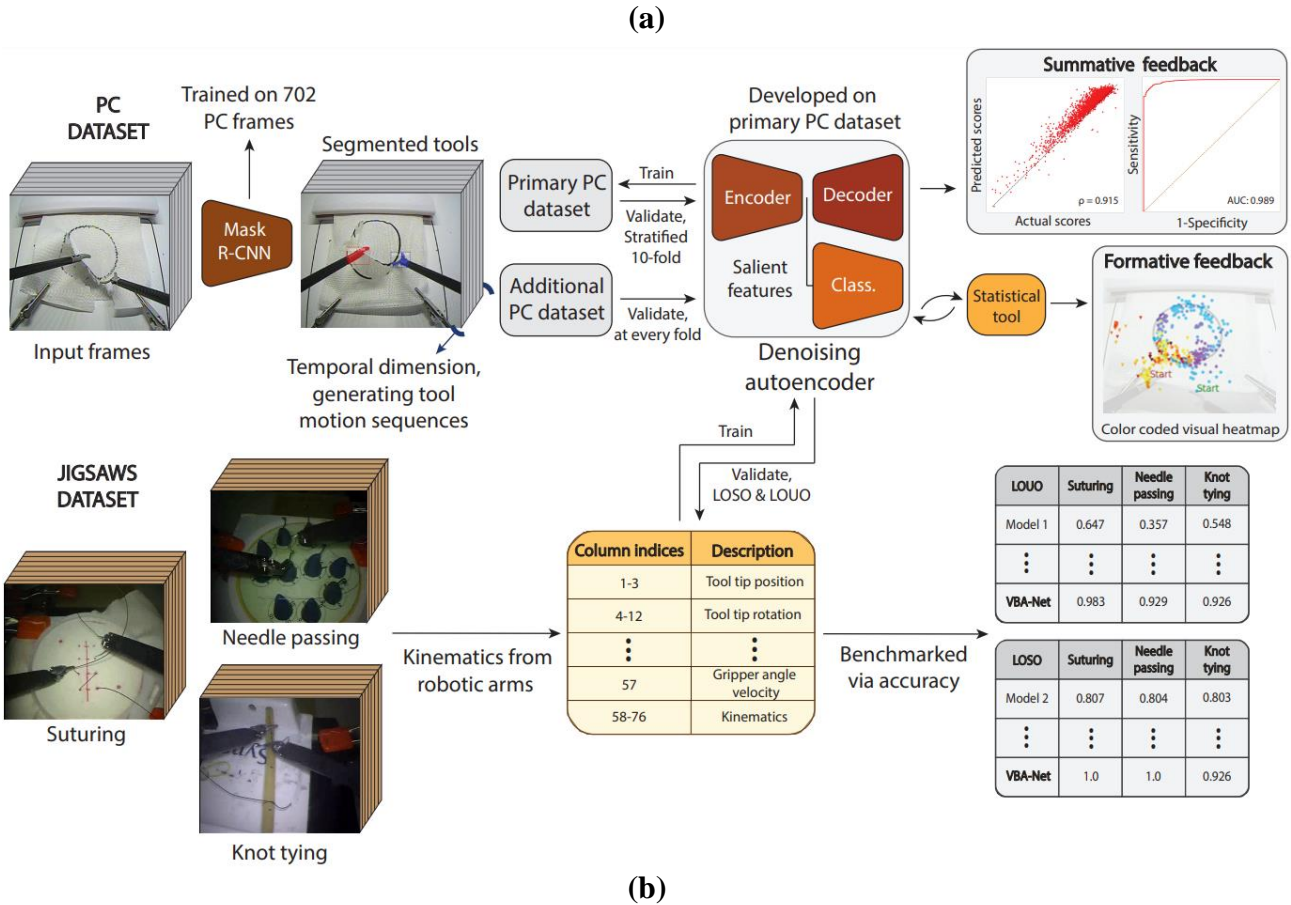

**(b)**

**Fig. 1 | Overview of the study. (a)** Subject demographics and descriptive data. **(b)** The pipeline of the VBA-Net. The model utilizes Mask R-CNN to generate tool motion sequences from video frames. Then denoising autoencoder (DAE) embeds the sequences for the classifier to predict summative and formative performance. The primary PC dataset is used to develop the model, i.e., tune its hyperparameters. The additional PC dataset, on the other hand, is used for validation. The JIGSAWS dataset is utilized to benchmark the model against the high-performing models in the literature.

Moreover, VBA-Net achieves an accuracy of 0.955±.002 while reporting 0.958 ± .003 and 0.922 ± .010 for sensitivity and specificity, respectively. Further, the model has an area under the curve (AUC) of 0.989±.001 for the receiver operating characteristics (ROC) curve (Fig. 2b).

The model's trustworthiness is analyzed on a single training session via trustworthiness metrics[21,22] (Supplementary information / Trustworthiness). Fig. 2c shows the trust spectrum accompanied by the NetTrustScore (NTS): a scalar overall score ranging 0-1 with 1 meaning highest trust, and the conditional NTS scores, i.e., NTS for the true and false predictions. The VBA-Net has robust trustworthiness with NTS values of 0.926 and 0.868 for the passing and failing classes. Moreover, for both the classes, the conditional NTS is above 0.9 when the prediction is true and around 0.3 when the prediction is false, implying that the VBA-Net has strong confidence in true predictions with low uncertainty while it can benefit from additional data for both the classes[22].

**Validation on the additional PC dataset.** For this analysis, we tested the VBA-Net, without retraining, on the additional PC dataset after every fold. This way, we could test the model's performance on the unseen subjects. As a result, the VBA-Net surpassed its performance on the primary PC dataset it was trained on and successfully predicted the FLS scores (Fig. 3a) with $\rho$ of 0.937 (with p <0.05 for every fold). In addition, for classification analysis, VBA-Net reported an accuracy of 0.876 ± .002, with sensitivity and specificity of 0.871 ± .005 and 0.887 ± .11, respectively. Finally, the VBA-Net's separability remained robust, with an AUC of 0.955 ± .002, as seen in Fig. 3b.

Fig. 3c shows the trust spectrum accompanied by the NTS and the conditional NTS scores. VBA-Net manages low uncertainty and high trustworthiness reporting NTS values of 0.844 and 0.831 for the passing and failing classes. When Fig. 3c is compared with Fig. 2c, we see the VBA-Net retains its prediction confidence for true predictions, while for passing cases, it reports lower NTS for false predictions, indicating the need for additional data on passing cases for the additional PC dataset. This is expected as the additional PC dataset has 202 passing samples compared to 1842 for the primary PC dataset (Extended Data Table 1).

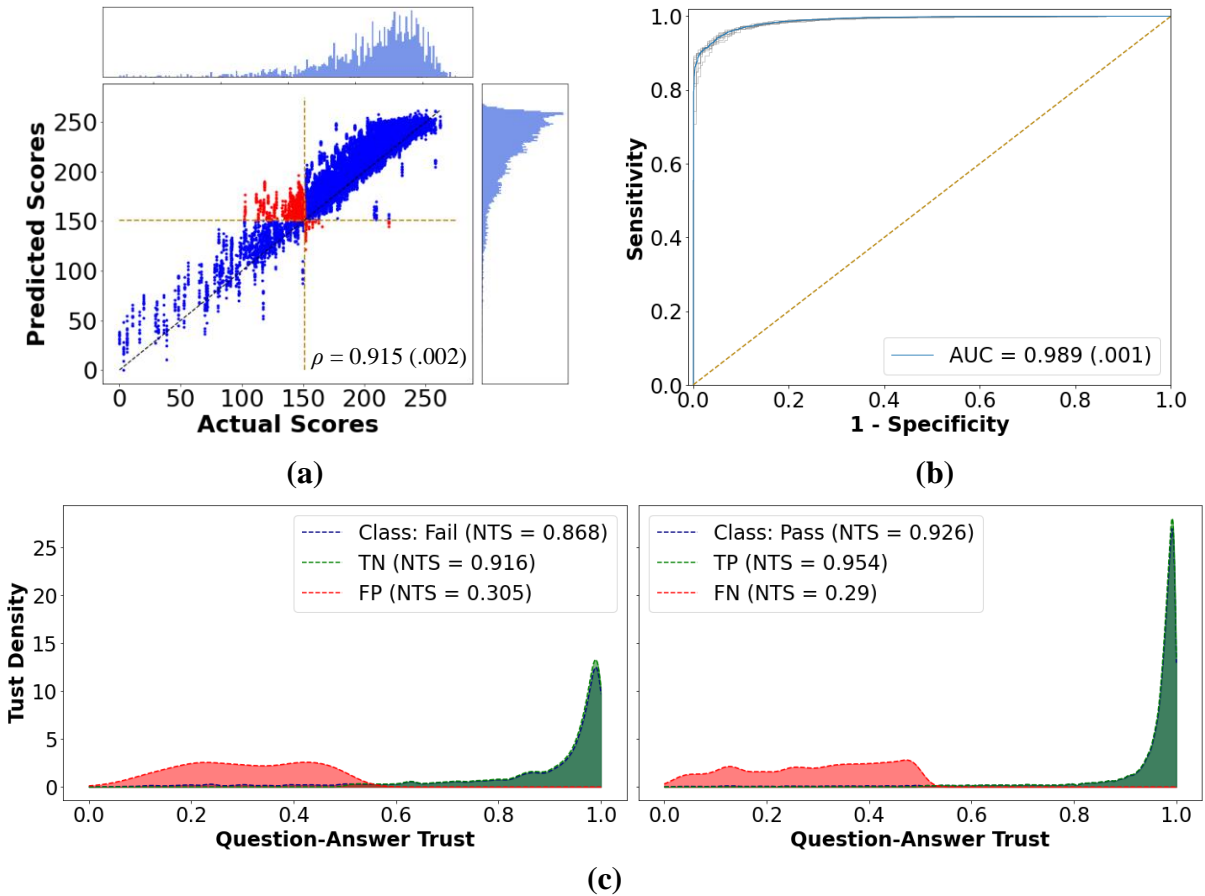

(a)

(b)

(c)

**Fig. 2 | Results for the primary PC datasets. (a)** Actual vs. predicted FLS scores for all ten training sessions. Here, the histograms show the frequency of samples for a given score. **(b)** The ROC curves. The yellow line represents the random chances. **(c)** Question-Answer trust plots for each class.

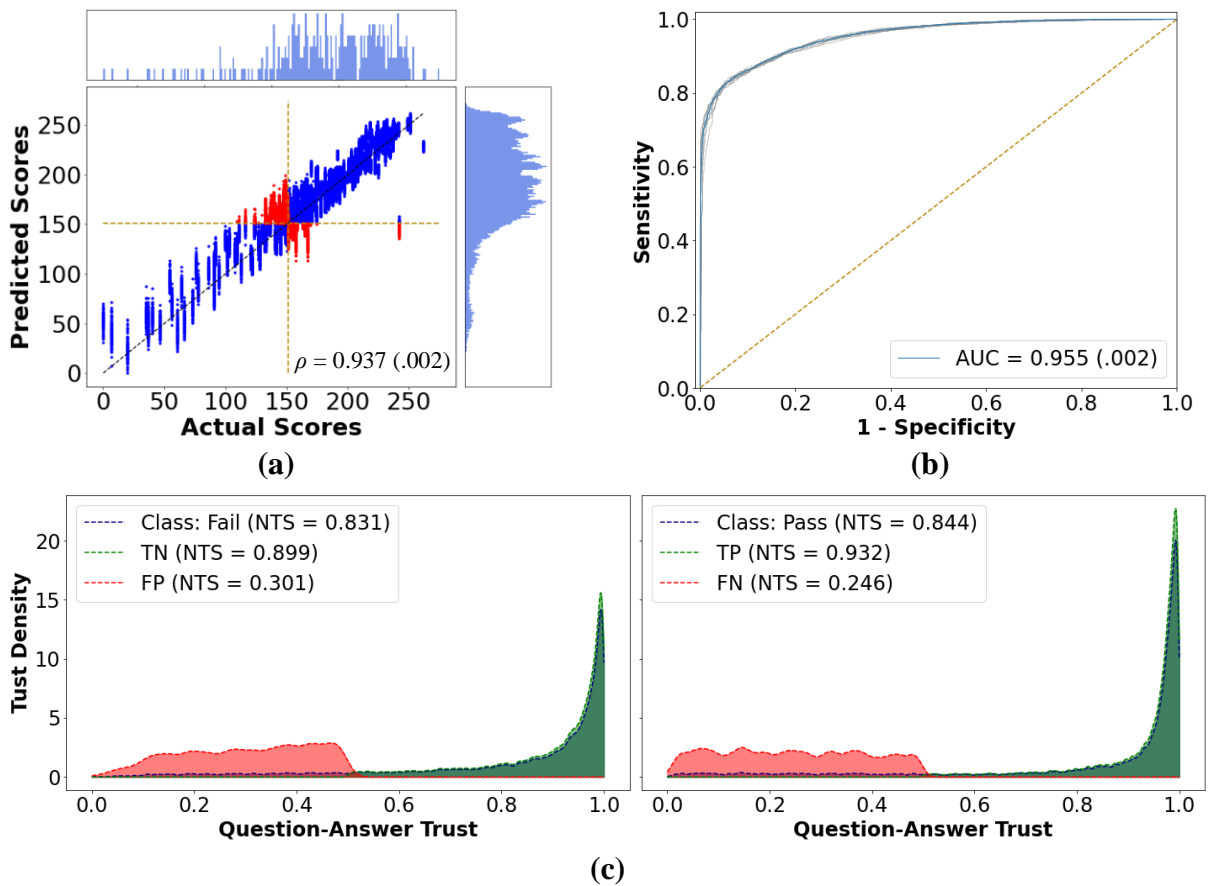

**Fig. 3 | Results for the additional PC datasets. (a)** Actual vs. predicted FLS scores for all ten runs. **(b)** The ROC curves. **(c)** Question-Answer trust plots for each class.

**Validation on the JIGSAWS dataset.** The generalizability of the VBA-Net on a separate task is demonstrated via leave-one-super-trial-out (LOSO) and leave-one-user-out (LOUO) CV schemes. Based upon the LOUO CV scheme, the VBA-Net outperformed the current state-of-the-art results in all three surgical tasks, reaching the highest overall average classification performance (accuracy= 0.946; Table 1). Notably, VBA-Net realized perfect accuracy for experts for all the tasks in the JIGSAWS dataset while misclassifying only two novice trials: one for suturing and one for knot tying (Extended Data Fig. 2a).

In addition, VBA-Net reported the highest Spearman correlation coefficients for both OSATS and GRS prediction for all the tasks (Table 2), achieving robust correlation for needle passing and knot tying while a moderate correlation for suturing. (For the breakdown of $\rho$, see Extended Data Table 2).

Table 3 presents the LOSO CV results and corresponding benchmark models that report at least 0.97 mean accuracy. (See Extended Data Table 3 for results <0.97). VBA-Net achieved perfect accuracy of 1.0 for suturing and needle passing tasks and provided an accuracy of 0.926 for the knot tying task, with a mean accuracy of

0.975, outperforming all the DL models (Extended Data Fig. 2b). Here, Fawaz *et al.* and Castro *et al.* were not included in the analysis because, in their LOSO scheme, they further divided the training set into train and validation without providing the split ratio. This is different from the standard LOSO protocol[14]. Likewise, Soleymani *et al.*[23] was excluded as they utilized a 4-fold (accuracy = 0.942) and 10-fold CV (accuracy = 0.973), respectively. Notably, a machine learning (ML) model[24] produces better mean accuracy than the VBA-Net. However, their approach is manually-intensive and not generalizable to other tasks.

Moreover, the mean $\rho_{OSATS}$ and $\rho_{GRS}$ were 0.63 and 0.77 for the LOSO CV, exceeding the state-of-the-art performance (Table 2). Other than suturing, where VBA-Net and Fawaz *et al.*[11] both reported 0.6, the VBA-Net outperformed the existing models in OSATS score prediction on all three tasks. For GRS prediction, on the other hand, the VBA-Net achieved the highest performance for each task. (For the breakdown of $\rho$, see Extended Data Table 4.)

**Table 1 | Classification scores for LOUO CV.**

| Author | Method | Suturing | Needle Passing | Knot Tying | Mean |
|---|---|---|---|---|---|
| Zia *et al.*[24] | DCT | 0.647 | 0.357 | 0.548 | 0.517 |
| Zia *et al.*[24] | DFT | 0.647 | 0.464 | 0.516 | 0.542 |
| Fard *et al.*[25] | Manual Features - LR | 0.705 | - | - | - |
| Fard *et al.*[25] | Manual Features - SVM | 0.721 | - | - | - |
| Funke *et al.*[9] | 3DCNN | - | - | 0.630 | - |
| Zia *et al.*[24] | ApEn | 0.882 | 0.857 | 0.774 | 0.838 |
| Khalid *et al.*[26] | Autoencoder | 0.840 | 0.840 | 0.840 | 0.840 |
| ***VBA-Net*** | ***DAE + Classifier*** | ***0.983*** | ***0.929*** | ***0.926*** | ***0.946*** |

DCT: Discrete Cosine Transform, DFT: Discrete Fourier Transform, LR: Logistic Regression, SVM: Support Vector Machine. ApEn: Approximate Entropy.

**Table 2 | Regression scores for LOSO & LOUO CV.** At each cell, the format follows $\rho_{OSATS}|\rho_{GRS}$.

| Author | Suturing | | Needle Passing | | Knot Tying | | Mean | |
|---|---|---|---|---|---|---|---|---|
| | LOSO | LOUO | LOSO | LOUO | LOSO | LOUO | LOSO | LOUO |
| Zia *et al.*[24] | 0.59\|0.75 | 0.45\|0.42 | 0.45\|0.53 | 0.53\|0.28* | 0.66\|0.76 | 0.56\|0.78 | 0.56\|0.68 | 0.51\|0.49 |
| Fawaz *et al.*[11] | 0.60\|- | n/a | 0.57\|- | n/a | 0.65\|- | n/a | 0.61\|- | n/a |
| ***VBA-Net*** | ***0.60\|0.76*** | ***0.52\|0.49*** | ***0.60\|0.73*** | ***0.74\|0.75*** | ***0.69\|0.83*** | ***0.80\|0.83*** | ***0.63\|0.77*** | ***0.69\|0.69*** |

*p > 0.05.

**Table 3 | Classification scores (>0.97) for LOSO CV and other CV schemes.**

| Author | Method | Suturing | Needle Passing | Knot Tying | Mean |
|---|---|---|---|---|---|

| Khalid et al.[26] | Autoencoder | 0.970 | 0.970 | 0.970 | 0.970 |
|---|---|---|---|---|---|
| Nguyen et al.[27] | CNN-LSTM | 0.984 | 0.984 | 0.948 | 0.972 |
| Fawaz et al.* [11,28] | CNN (Adjusted LOSO) | 1.0 | 1.0 | 0.921 | 0.974 |
| Funke et al.[9] | 3DCNN + TSN | 1.0 | 0.964 | 0.958 | 0.974 |
| Soleymani et al. * [23] | CNN + FFT (10-fold) | N/A | N/A | N/A | 0.973 |
| **VBA-Net** | **DAE + Classifier** | **1.0** | **1.0** | **0.926** | **0.975** |
| Castro et al.* [12] | CNN (Adjusted LOSO) | 0.984 | 0.989 | 0.989 | 0.987 |
| Zia et al.[24] | ApEn** | 1.0 | 1.0 | 0.999 | 0.999 |

TSN: temporal segment networks. FFT: Fast Fourier Transform. *Shown in the table but were not included in the comparison.

## Discussion

VBA has garnered significant attention for surgical skill assessment following the shift to competency-based medical education and patient safety. It promises to enhance the formative assessment of the learning process by offering trainees timely feedback while also allowing experienced surgeons to reflect on their surgical techniques. However, VBA methods need to be scalable, generalizable, and demonstrate a high level of correlation with current summative methods employed in the field. Herein, we demonstrated that VBA-Net offers excellent and trustworthy performances in various surgical procedures. The performance metrics presented in the previous section underscore the effectiveness of the VBA-Net in objective and automated summative score prediction.

VBA-Net can generalize well to unseen data. Thus, it can help individual trainees prepare for high-stakes certification exams such as FLS by providing reproducible scores in real-time. Moreover, VBA-Net generalizes well to unseen subjects. Hence, it can assist proctors with the certification process as each subject performs one-time and receives an end-point result. Besides, the model yields solid binary classification performance, particularly for specificity, i.e., the model was robust in detecting false certification for both unseen data and subjects. This finding is important as human error is one of the leading causes of death in the OR; hence poor clinical outcomes[29] and preventing false certification can greatly reduce that. In summary, these attributes can significantly contribute to more robust validity evidence, i.e., improving patient outcomes.

We validated the generalizability of the VBA-Net by benchmarking it against the state-of-the-art models on the JIGSAWS dataset. Based on the LOUO CV, the VBA-Net improved the average OSATS and GRS score predictions by 35.3% and 40.8%. Further, VBA-Net outperformed the closest ML model[24] with a 12.9% margin and the closest DL model[9] in knot tying with a 47% margin in classifying the surgeons. This shows that the VBA-Net can generalize to tasks other than PC and can predict OSATS scores of new subjects.

In addition, for unseen trials measured via LOSO, VBA-Net achieved the highest Spearman correlation coefficient in predicting both the OSATS and GRS scores, indicating that the VBA-net can predict the performance on the unseen trials better, supporting proctoring of the trainees. Here, the model reported comparatively lower accuracy in knot tying for classification analysis. We can attribute this to the complexity of the knot tying task as stated in the literature[9,28]. Besides, when comparing LOUO with LOSO, we observed a decrease in classification and regression performances, signifying that the subjects demonstrated class-specific bimanual motor behavior. Finally, we noticed that several studies[9,11,12,24,28], including ours, reported perfect accuracy on Suturing and Needle Passing tasks via LOSO. Therefore, we believe the field can benefit from new publicly available surgical datasets.

We now discuss how VBA-Net provides formative feedback via a *post-hoc* explainability tool, i.e., CAM (see Supplementary information / CAM for details). Fig. 4 shows the 2D CAMs projected onto the tool trajectory using a 1D color-coded contour for a true positive (pass) (Fig. 4a) and a true negative (fail) (Fig. 4b) case.

In Fig. 4a, we provide an example for a true negative case (fail) and annotate (red) the locations corresponding to poor performance based on the surgical videos. The subject started smoothly, successfully reaching the circular pattern from the corner of the gauze without unnecessary movement. However, they failed to cut the first half of the circle after multiple attempts and eventually moved to the second half without completing the first half. The high activation pointed by 'arrow 1' captures this behavior. Simultaneously, the grasper was repositioned from the lower-left corner of the gauze to the middle left, a move that was not observed in the passing cases. The subject struggled through the second half, failing to cut the circle while holding the gauze with the grasper. The high activation at 'arrow 2' captures this.

We also analyzed a true positive case (pass) in Fig. 4b. Here, we annotated the desired performance (green) based on the corresponding video. As a result, we observed that the subject uses the grasper effectively and avoids unnecessary grip attempts (arrow 3). Moreover, the subject uses smooth motion and cuts the gauze fast when cutting the second half of the circular pattern. This desired behavior is captured by the network as pointed by 'arrow 4'. Overall, we observed that the activations are independent of the duration and are specific to each trial.

To establish the effectiveness of such formative assessment without expert guidance, we analyzed CAMs via a model-agnostic statistical tool (see Supplementary information / Model agnostic statistical tool for CAM validation for details). We hypothesized that if CAMs highlight the salient parts, the model should distinguish better between skill classes when the input sequences are masked with CAM. Consequently, the results should improve. We implemented our approach on the primary PC dataset on a single training session with a stratified 10-fold CV. Resulting training we obtained distribution of metrics for each fold before and after masking. When compared the distribution's mean for each metric (Fig. 4c), after-masking case achieved significantly greater performance than before-masking case.

These results signify that the CAMs are valid and highlight the essential parts of the sequence towards the skill class. Hence they can be used for low-stakes informative feedback. Moreover, such visual maps can draw the proctor's attention to the distinct parts of the videos, thus improving the time-effectiveness of the assessment, i.e., it can reduce the workload and burnout. Moreover, these validated maps can lead to objective and automated editing to establish time-efficient and generalizable low-stakes rubrics for surgical education.

Still, our study has several limitations. First, tool trajectories are the only extracted features from the videos; hence, while the literature is well-established on tool tracking, whether it is the optimal feature set remains open. Second, our model is not end-to-end. This has its strength in using either videos or kinematics as inputs, but it increases the framework's complexity which could be reduced using an end-to-end model.

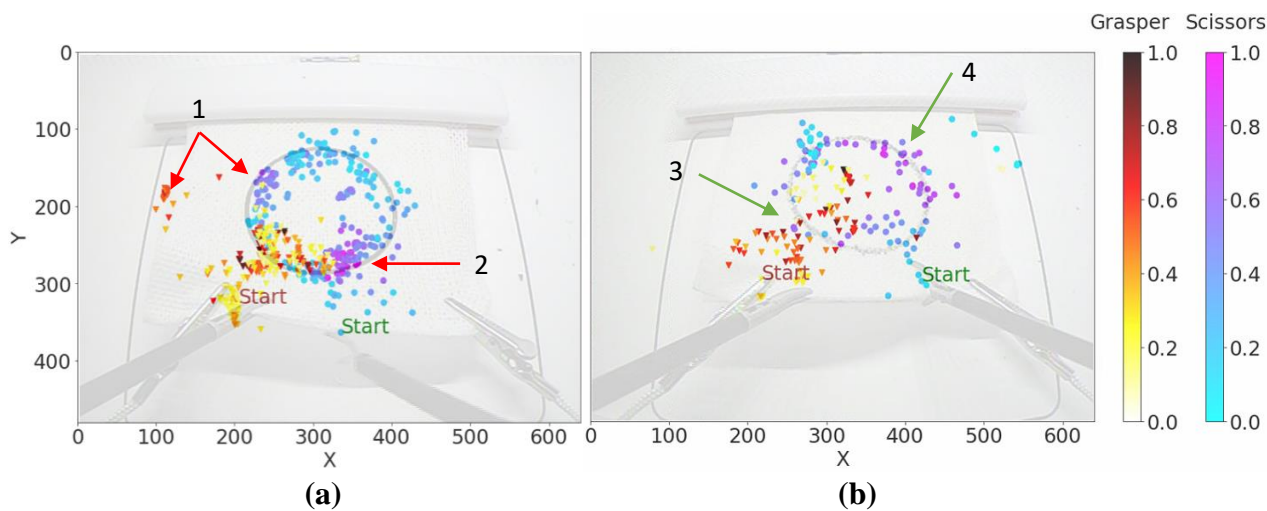

| Model | Accuracy (p < .05) | Sensitivity (p < .05) | Specificity (p < .05) |
|---|---|---|---|
| Before Masking | 0.951 (.019) | 0.954 (.019) | 0.925 (.056) |
| After Masking | 0.997 (.004) | 0.998 (.004) | 0.990 (.019) |

(c)

**Fig. 4 | CAM results.** CAM plots for **(a)** a true negative (FLS score: 16.8) and **(b)** a true positive (FLS score: 170.7) sample. The plots are presented in the original frame size of 640 x 480. The red arrows indicate tool motions that may lead to poor performance, while the green arrows indicate smooth behavior. The color-coded heatmaps illustrate the intensities of the same CAM generated for the given samples. However, different color maps are used for scissors and grasper locations. **(c)** Overall VBA-Net performance comparison before and after masking. Here, p is the p-value of the statistical analysis, and the numbers within the parentheses in the second and third rows represent standard deviation based on 10-folds of training.

## Methods

**Development and validation datasets.** The PC videos were collected via the standard FLS box camera with 640 x 480 resolution at 30 FPS. Before the PC trials, all 33 medical students were consented and were informed of the experimental protocol, approved by the Institutional Review Board (IRB) of the University at Buffalo and Rensselaer Polytechnic Institute (RPI).

For the development dataset, primary PC, 21 subjects participated for 12 days. On the first day, each subject executed the task once. Between days 2 and 12, subjects performed up to ten PC trials. Finally, on the last day, five repetitions were reported by each subject, totaling around 99 trials per subject in the dataset, after the ones with negative FLS scores were removed. Notably, this resulted in an imbalanced dataset where the pass/fail ratio is 8.9. For the validation dataset, additional PC dataset, 12 subjects were asked to perform up to 26 times in one day. We did not observe an imbalance in this dataset.

**Model development.** Several studies have shown the effectiveness of video-based instrument tracking and detection towards objective and automated assessment of skills[30–35]. Therefore, we developed a video-based DL pipeline in which the surgical videos are processed to gather tool motion sequences from videos for skill assessment. For this, we used an instance segmentation model, namely Mask R-CNN (For the architecture see Extended Data Fig. 3). Instance segmentation differs from object detection as the background is also a class in training and the model learns to segment the instance out of its background. This is beneficial when working on video-based datasets with a constant camera view, e.g., the PC datasets, especially when background items such as clips resembling the surgical tools in use. Once trained, Mask R-CNN was used in each trial to generate motion sequences. The coordinates of the succeeding and preceding frames were averaged for frames in which the model failed to detect the tools.

We extracted embedded features of the tool motion sequences via a denoising autoencoder (DAE) (For the architecture see Extended Data Fig. 4) with Gaussian noise (alpha = 0.001) before processing them for skill assessment. The DAE is an unsupervised CNN-based autoencoder used in surgical skills assessment in several studies[26,30,35,36]. Finally, we utilized a CNN-based classifier (Extended Data Fig. 4) on the salient motion sequences to assess summative skills and provide formative feedback (For the architecture, training and testing details and used metrics, see Supplementary information / Model structure details, Model training & Model testing and Evaluation metrics).

## Acknowledgments

The authors gratefully acknowledge the support of this work through the Medical Technology Enterprise Consortium (MTEC) award #W81XWH2090019 (2020-628) and the U.S. Army Futures Command, Combat Capabilities Development Command Soldier Center STTC cooperative research agreement #W911NF-17-2-0022.

## Author contributions

E.Y., X.I., and S.D. conceived the idea. E.Y. collected the data, annotated the videos, developed the model pipeline, conducted the statistical and data analysis, and drafted the manuscript. U.K. was involved with the statistical analyses. R.R. advised the use of denoising autoencoder. U.K., X.I., and S.D. were responsible for supervision and revision of the intellectual content in the manuscript. S.D. was responsible for funding acquisition.

## Competing interests

The authors declare no competing interests.

## Extended Data - Figures

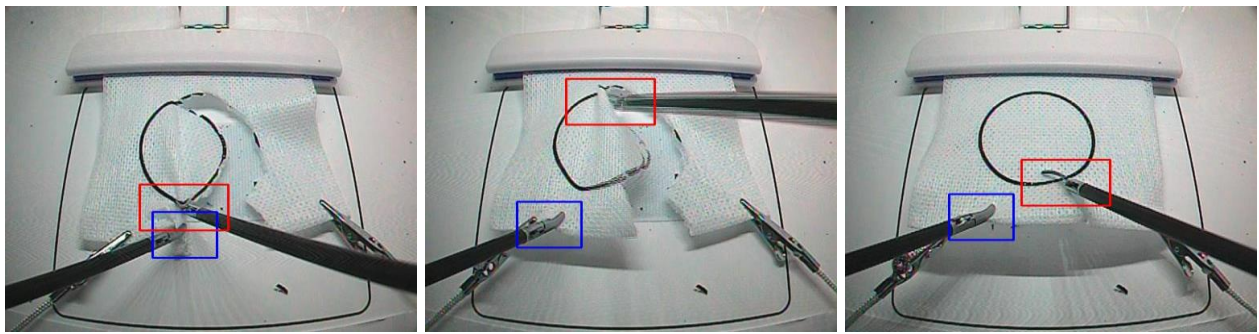

**Extended Data Fig. 1 | Mask R-CNN output.** Sample regressed bounding boxes of the tools. It is observed that the Mask R-CNN can successfully locate the tools when there is an intersection (left), blurry image (middle), and different tool angles (right).

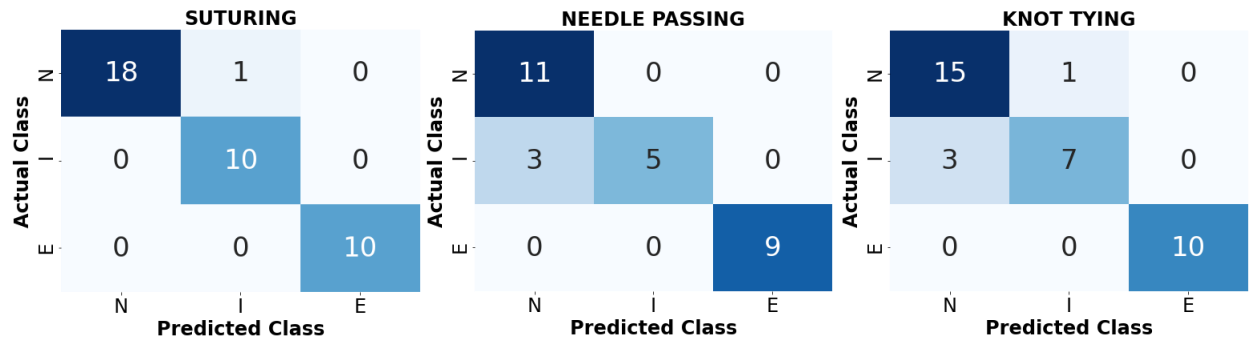

**(a)**

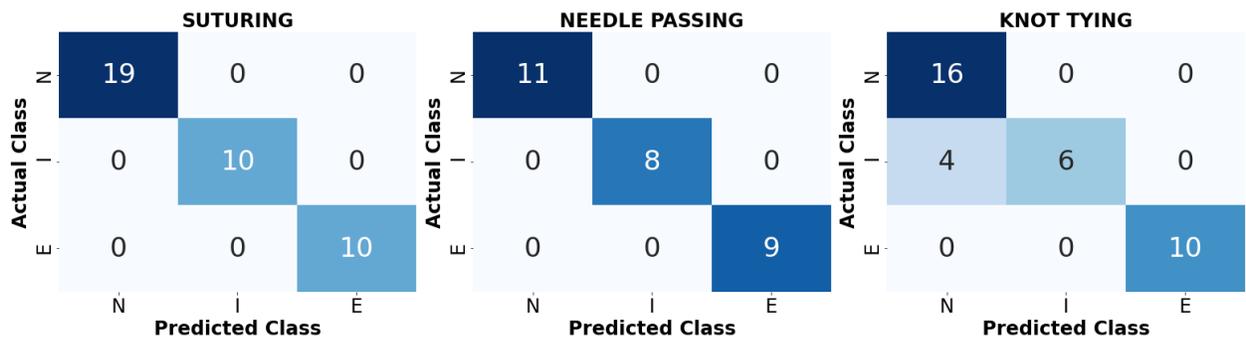

**(b)**

**Extended Data Fig. 2 | Confusion matrices for the surgical datasets. (a)** For the JIGSAWS dataset via the LOUO CV scheme. Here, N, I, and E stand for Novice, Intermediate, and Expert, respectively. **(b)** For the JIGSAWS dataset via the LOSO CV scheme.

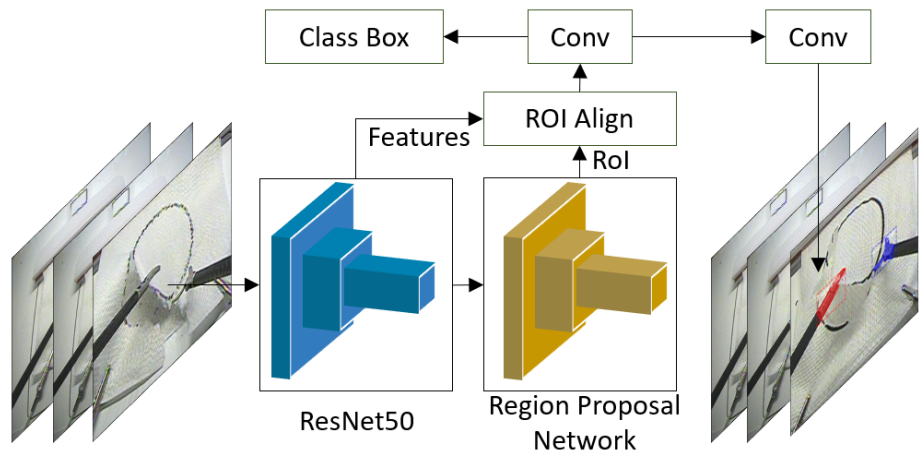

**Extended Data Fig. 3 | Mask R-CNN architecture.** It consists of a CNN-backbone, ResNet50, and Region Proposal Network (RPN) to output object properties.

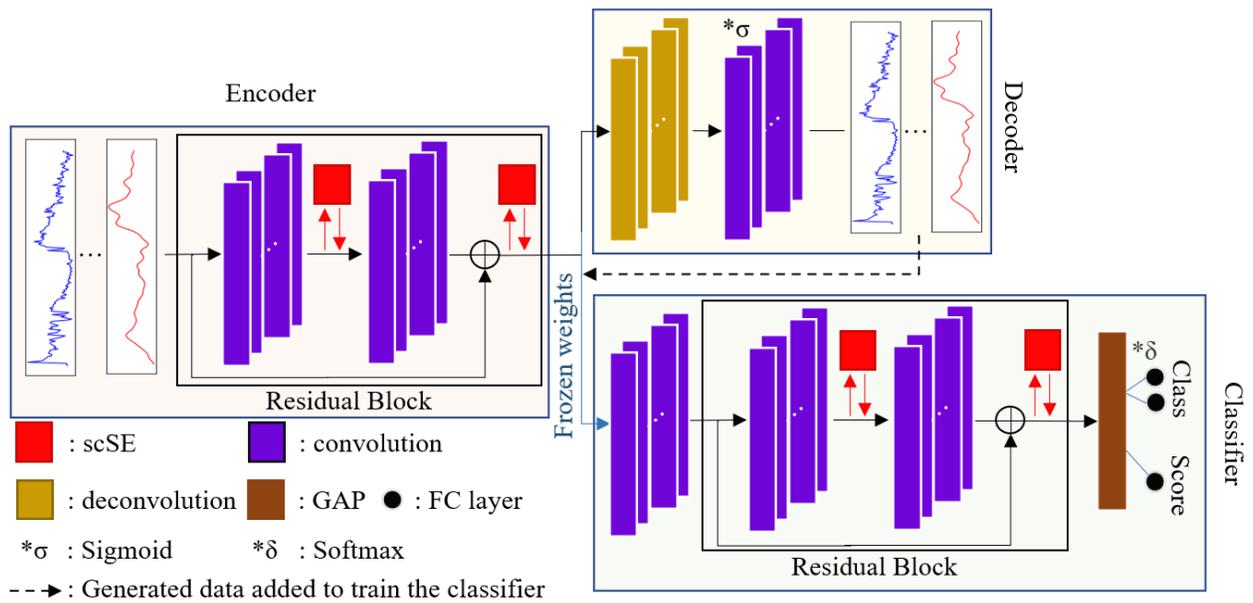

**Extended Data Fig. 4 | The DAE and classifier structure.** The figure illustrates the encoder, decoder, and classifier in their utilized order.

## Extended Data - Tables

**Extended Data Table 1 | FLS score statistics based on binary classes for the PC datasets.**

| Dataset | | No. of samples | Mean Duration (sec) | STD Duration (sec) | Mean FLS score | STD FLS score |
|---|---|---|---|---|---|---|
| **Main** | **Pass** | 1842 | 73.5 | 25.0 | 208.3 | 24.3 |
| | **Fail** | 213 | 161.9 | 36.9 | 115.3 | 33.7 |
| | **Overall** | 2,055 | 82.7 | 37.8 | 198.7 | 38.1 |
| **Additional** | **Pass** | 202 | 86.0 | 23.4 | 185.2 | 19.3 |
| | **Fail** | 105 | 148.5 | 33.0 | 121.8 | 31.0 |
| | **Overall** | 307 | 107.4 | 40.2 | 163.5 | 38.5 |

**Extended Data Table 2 | Breakdown of Spearman correlation coefficients, $\rho$s, for OSATS scores prediction via the LOUO CV.**

| Task | Respect for tissue | Suture/needle handling | Time and motion | Flow of operation | Overall performance | Quality of final product |
|---|---|---|---|---|---|---|
| **ST** | 0.25* | 0.67 | 0.66 | 0.59 | 0.53 | 0.43 |
| **NP** | 0.69 | 0.70 | 0.89 | 0.76 | 0.72 | 0.68 |
| **KT** | 0.75 | 0.78 | 0.86 | 0.78 | 0.76 | 0.87 |

*$p > 0.05$. ST: suturing, NP: needle passing, KT: knot tying.

**Extended Data Table 3 | Classification scores (<0.97) for LOSO CV and other CV schemes.**

| Author | Method | Suturing | Needle Passing | Knot Tying | Mean |
|---|---|---|---|---|---|
| Lajkó et al.[35] | CNN | 0.807 | 0.797 | 0.804 | 0.803 |
| Anh et al.[36] | Autoencoder | 0.835 | 0.823 | 0.806 | 0.821 |
| Lajkó et al.[35] | CNN + LSTM | 0.816 | 0.832 | 0.828 | 0.825 |
| Lajkó et al.[35] | ResNet | 0.819 | 0.842 | 0.835 | 0.832 |
| Anh et al.[36] | LSTM | 0.951 | 0.915 | 0.896 | 0.921 |
| Wang et al.[37] | CNN | 0.925 | 0.954 | 0.913 | 0.931 |
| Anh et al.[36] | CNN-LSTM | 0.964 | 0.934 | 0.910 | 0.936 |
| Anh et al.[36] | LSTM | 0.965 | 0.941 | 0.912 | 0.940 |
| Soleymani et al.*[23] | CNN + FFT (4-fold) | N/A | N/A | N/A | 0.942 |
| Anh et al.[36] | CNN | 0.968 | 0.954 | 0.927 | 0.950 |
| Tao et al.[38] | Sparse-HMM | 0.974 | 0.962 | 0.944 | 0.960 |
| Wang et al [39] | CNN-GRU | N/A | N/A | N/A | 0.960 |
| *VBA-Net* | *DAE* | *1.0* | *1.0* | *0.926* | *0.975* |

**Extended Data Table 4 | Breakdown of Spearman correlation coefficients, $\rho$s, for OSATS score prediction via the LOSO CV.**

| Task | Respect for tissue | Suture/needle handling | Time and motion | Flow of operation | Overall performance | Quality of final product |
|---|---|---|---|---|---|---|
| **ST** | 0.51 | 0.63 | 0.64 | 0.57 | 0.58 | 0.68 |
| **NP** | 0.52 | 0.77 | 0.65 | 0.69 | 0.55 | 0.43 |
| **KT** | 0.68 | 0.65 | 0.76 | 0.59 | 0.80 | 0.64 |

ST: suturing, NP: needle passing, KT: knot tying.

## Supplementary information

**Trustworthiness.** Besides the commonly used metrics, we utilize the recently proposed trustworthiness metrics[22,40], i.e., question-answer trust, trust density, conditional trust density, trust spectrum, and NetTrustScore (NTS), to assess the reliability of the VBA-Net on the classification results. In this concept, the Softmax probability is associated with confidence, C(y|x), and a model, *M*, is trustworthy when a true prediction is accompanied by stronger Softmax and vice versa. Eqn. 1 presents the question-answer trust.

$$Q_z(x,y) = \begin{cases} C(y|x)^\alpha & \text{if } x \in R_{y=z}|M \\ 1 - C(y|x)^\beta & \text{if } x \in R_{y \neq z}|M \end{cases}$$

In Eqn. 1, $R_{y=z}$ is the space of all the samples (x) for which the predicted ($y$) and the actual ($z$) classes match. On the other hand, $R_{y\neq z}$ is the space where they do not. Next, $\alpha$ rewards confidence for true predictions and $\beta$ penalizes over-confidence when the forecast is incorrect. In this study, both are set to 1. Finally, $Q_z(x, y)$ denotes the question-answer trust for a given class ($z$).

Next, trust density is the probability density distribution of $Q_z(x, y)$ mapped via the non-parametric density estimation with a Gaussian kernel [40]. Moreover, conditional trust density takes trust density one step further by calculating the distributions separately for when $R_{y=z}$ and $R_{y\neq z}$. It helps spot overconfidence and overcaution for a given class (z). As a remark, in binary classification, $R_{y=z}$ represents the TP or TN whereas $R_{y\neq z}$ represents the FN or FP depending on the selected attribute, i.e., positive & negative.

The trust spectrum, $T_M(z)$, represents the overall trust behavior based on every class and NTS ($T_M$) is the overall trustworthiness score generated by integrating the trust spectrum, see Eqn. 2.

$$T_M(z) = \frac{1}{N} \int Q_z(x)dx, \qquad NTS\ (T_M) = \int P(z)T_M(z)dz \qquad \text{)}$$

Here, $N$ is the sample size for a given class.

**Class Activation Map (CAM).** CAM is a visualization tool highlighting the regions that contribute most to the classification prediction. It is based on the Hadamard product of the pre-Softmax weights and the activations of the last convolution[13]. If $f_k(i)$ represents the activations at the convolutional layer preceding global average pooling (GAP) for the unit k and timestamp $i$ and $w_k{}^c$ is the pre-softmax weights between the GAP layer and the fully-connected classifier for the same unit and class $c$, CAM is defined as $M_c(i) = \sum_k w_k{}^c f_k(i)$.

**Model-agnostic statistical tool for CAM validation.** First, we mask each input by element-wise multiplying them with their respective CAMs. Then we train the VBA-Net again from scratch using the weighted inputs and evaluate it via the stratified 10-fold CV. As a result, we end with two distributions for the given metrics, e.g., accuracy: before-masking and after-masking, both of which has ten samples reflecting the selected CV scheme. Next, we employ one-sided Wilcoxon sign test to check whether the mean of the distribution is significantly different for the after-masking scenario. Here, our null hypothesis, $H_0$, presumes no significant difference, whereas the alternative hypothesis, $H_1$, assumes that the mean of the distribution for the after-masking is significantly greater. The significance is 0.05 for this analysis.

**Model architecture.** Mask R-CNN[17] first extracts spatial features from the input frames using a CNN backbone, i.e., ResNet50, to avoid the risk of overfitting. The spatial features are then processed in

Region Proposal Network (RPN), generating regions of interest (RoI) for each instance. Here, RoI is assumed correct for detection confidences of 0.7 or higher. Next, RoIPool is applied to a third of the RoI to extract salient feature maps, and the RoIAlign algorithm is imposed to align the pre and post-RPN features. Finally, the generated features are fed into the convolutional layers, outputting the class and the binary mask for each instance and the respective bounding box coordinates.

The second stage of the VBA-Net is the DAE and the classifier. DAE consists of an encoder to extract the important features from the input and a decoder to reconstruct the input based on the features provided by the encoder. Finally, the classifier utilizes the encoder, with its weights frozen after training to assess skills.

We utilized an in-house attention-infused residual block to prevent the vanishing gradient problem[41] for both the encoder and the classifier. Specifically, our residual block consisted of two identical convolutional layers and an identity layer. Moreover, two spatial and channel squeeze and channel excitation (scSE)[42] attention layers were included for their ability to recalibrate the input feature maps by highlighting the most salient features in the residual block. The first scSE was placed between the initial and second convolutional layers. The second scSE was after the residual weights were added to the second convolutional layer. In addition, the convolutional layers within the residual block were dilated when training for classifier[43].

When training the classifier, a Global Average Pooling (GAP)[44] layer followed the residual block, aggregating the feature maps and feeding them to the fully connected layer while allowing training the model with inputs of different sizes. Lastly, a fully connected layer consisting of one node and no activation when trained for regression and two nodes and Softmax activation when trained for binary classification were added to output the FLS scores and skill classes, respectively.

**Model training.** We pre-train Mask R-CNN on the COCO dataset[20] and fine-tune the classifier on the pattern cutting (PC) frames from both the PC datasets. Further, the output layer is configured to accommodate each class, i.e., scissors, grasper, and the background. 702 frames are randomly selected for training from all 2362 videos where both scissors and grasper are available. This is to optimize the coverage of conflicting scenarios during training. 562 (80%) frames are used to train and validate Mask R-CNN and 140 (20%) for testing. Among these 562 frames, 450 (80%) and 112 (20%) frames are used for training and validation, respectively. Further, all the frames are resized to 512 x 512 from 640 x 480 for Mask R-CNN. Finally, the VGG Image Annotator (VIA)[45] is used to annotate scissors and grasper tooltips in each frame using polygon annotation, the standard input for Mask R-CNN[17]

We train only the heads of the Mask Region-based Convolutional Neural Network (Mask R-CNN) for 40 epochs while keeping the remaining layers frozen. During training, the learning rate is initiated as 0.001, and the weight decay and momentum are 0.0003 and 0.9, respectively. The learning rate is decreased by a factor of ten after the $20^{th}$ epoch. Finally, we augment the frames by implementing Gaussian Blur (sigma = 0-5) and horizontal flipping 50% of the time, per epoch.

Before training the denoising autoencoder (DAE) and the classifier via the extracted motion sequences, each sequence is downsampled to 1 FPS to reduce training time[33]. Moreover, the sequences are normalized using min-max normalization. Lastly, the performance scores are pre-processed via z-normalization, and one hot encoding was used for the class labels. When validating the model, kinematics of the JIGSAWS dataset are pre-processed through the same pipeline as the PC datasets.

To extract salient features from motion sequences via the DAE, binary cross-entropy is minimized with a learning rate of 0.001. On the other hand, the classifier is trained with a mean squared error loss function when predicting the FLS scores and cosine similarity when performing classification, as it is shown to provide superior results on datasets with a limited sample size when trained from scratch[46]. The learning rate is 0.0002. Further, an Adam optimizer is used when training the model. Finally, the Scaled Exponential Linear Unit (SELU)[12] activation function is used for all the convolutional layers unless stated otherwise in Extended Data Fig. 4.

The batch size is one during training because each input has a different sequential length. The training is regulated using early-stopping based-on validation loss with the patience of 4 and 20 epochs for denoising autoencoder and classifier training, respectively, for the PC datasets. These values are 40 and 200 when the JIGSAWS dataset[14] is utilized. Further, L2 regularizing (0.00001) is applied as both the kernel and activity regularizer[12]. Moreover, we incorporate class weights into the training to account for imbalance. Finally, the training is conducted on a workstation with AMD Ryzen 7 2700X and NVIDIA GeForce RTX 2070.

**Model testing.** When evaluating the VBA-Net on the PC dataset, Mask R-CNN is tested via train/validation/test split CV. On the other hand, the DAE and classifier are tested via a stratified 10-fold CV utilizing all the available data while preserving the imbalance ratio in each fold.

To evaluate the benchmarked models, we employ the standard CV schemes for JIGSAWS, i.e., leave-one-supertrial-out (LOSO) and leave-one-user-out (LOUO). LOSO CV scheme is a specialized version of the k-fold CV used by the majority of the papers on the JIGSAWS dataset. In LOSO, the i[th] trial of each participant is used for testing, while the remaining trials are used for training the network. Thus, LOSO is advantageous in assessing the model's performance on unseen data. However, it is specifically developed for the JIGSAWS dataset and has limited utility in the literature. Furthermore, LOSO is not informative for the cases where the model evaluates new surgeons. LOUO overcomes this limitation. In the LOUO CV scheme, the trials of a single subject are removed from the training process and used to test the model. This is repeated for each subject. Therefore, the network is challenged to generalize to an unseen subject. Moreover, LOUO can be used with any dataset with more than one subject performing. The downside of LOUO is that it is blind to the model's performance on unseen data of the same subject, a crucial element for training.

**Evaluation metrics.** Mask R-CNN is evaluated via average precision with intersection over union (IoU) being 0.5 to consider the predicted bounding box true[17,20]. We employ Spearman correlation coefficient ($\rho$) to evaluate the score prediction performance, whereas accuracy, sensitivity, specificity, and area under curve (AUC) of the Receiver Operating Characteristics (ROC) curve are used to assess the binary classification results. When benchmarking the VBA-Net on the multi-class JIGSAWS dataset, we employ accuracy to evaluate the classification results. In contrast, $\rho$ is used for OSATS ($\rho_{OSATS}$) and GRS ($\rho_{GRS}$) score predictions where $\rho_{OSATS}$ is the mean value of $\rho$s for every six OSATS subscores detailed in Tables S2,3[11,14,24].